\documentclass{Interspeech2024}

\interspeechcameraready 

\title{How to Evaluate Automatic Speech Recognition: Comparing Different Performance and Bias Measures}

\name[affiliation={1}]{Tanvina}{Patel}
\name[affiliation={1,2}]{Wiebke}{Hutiri}
\name[affiliation={1}]{Aaron Yi}{Ding}
\name[affiliation={1}]{Odette}{Scharenborg}

\address{
  $^1$Delft University of Technology, The Netherlands\\
  $^2$Sony AI, Switzerland  }
\email{t.b.patel@tudelft.nl, wiebke.hutiri@sony.com, aaron.ding@tudelft.nl, o.e.scharenborg@tudelft.nl}


\usepackage{colortbl}
\usepackage{comment}
\usepackage{booktabs}
\usepackage{multirow}
\usepackage{hyperref}

\begin{document}
\maketitle

\begin{abstract}

There is increasingly more evidence that automatic speech recognition (ASR) systems are biased against different speakers and speaker groups, e.g., due to gender, age, or accent. Research on bias in ASR has so far primarily focused on detecting and quantifying bias, and developing mitigation approaches. Despite this progress, the open question is how to measure the performance and bias of a system. In this study, we compare different performance and bias measures, from literature and proposed, to evaluate state-of-the-art end-to-end ASR systems for Dutch. Our experiments use several bias mitigation strategies to address bias against different speaker groups. The findings reveal that averaged error rates, a standard in ASR research, alone is not sufficient and should be supplemented by other measures. The paper ends with recommendations for reporting ASR performance and bias to better represent a system’s performance for diverse speaker groups, and overall system bias.
    
\end{abstract}

\section{Introduction}
Automatic speech processing technologies enhance digital inclusion with user-friendly interfaces, ensuring accessibility regardless of sight, mobility, or literacy~\cite{Seaborn2021voice}. Large datasets and advances in deep learning have significantly improved the performance of speech technologies~\cite{ArticleSpeechTech,MEHRISH2023101869}. Despite improved performance observed for standard or `norm' speakers on different speech-related tasks~\cite{yang21c_s_superb}, challenges persist, resulting in sub-optimal performance for many `diverse' users~\cite{FENG2024_101567, AccentGapWashingtonPost, koenecke2020racial}, e.g., due to  age, language, accent, health, country of origin (see Sec.~2.1 for a review). In machine learning (ML), these performance differences produced by a model for specific speaker groups based on personal attributes is referred to as \textit{bias}. There is also increasingly more evidence of unequal performance and bias in commercial speech recognition systems, due to gender~\cite{RachaelTatman_Genderbias}, accents~\cite{AccentGapWashingtonPost}, race~\cite{HarvardReview_race_genderbais}, and underrepresented groups~\cite{ScienticAmerican}. These findings together highlight the need to study bias in Automatic Speech Recognition (ASR). Performance differences in ASR systems can stem from different known and unknown sources: skewed training data and cultural nuances in the training data, inconsistent human transcriptions, type of neural network architecture used, the already mentioned speaker-dependent characteristics (age, speaking style, accent, emotion, and short-term health conditions)~\cite{FENG2024_101567, Dheram2022TowardDisparities}, and speaker-independent factors (environment, recording technology, and application setting)~\cite{Noise_ASR_TASLP}. 
 
Quantifying and addressing performance differences is crucial for mitigating bias and promoting fairness in speech recognition. 
\textit{Bias measures} (or fairness criteria) that quantify and measure bias are calculated from statistical \textit{base metrics}. In ASR, the primary base metric used is Word Error Rate (WER), Phone Error Rate (PER), and Character Error Rate (CER), with occasional modifications such as excluding deletions~\cite{Tatman2017EffectsCaptions},  perplexity to evaluate language models~\cite{koenecke2020racial}, and confidence scores to estimate the probability of an output without relying solely on ground truth transcriptions~\cite{Dheram2022TowardDisparities}. Bias measures typically measure bias between different speaker groups~\cite{Mehrabi2019Survey}. In ASR, bias is generally a difference or disparity in error rates between speaker groups \cite{Dheram2022TowardDisparities, feng2021quantifying}.
Most bias measures in the literature have been developed for binary decision outcomes and binary groups (i.e. one group with a protected attribute and an unprotected group). This makes model evaluation and selection extremely difficult if bias must be evaluated between many speaker groups, e.g., different age or accent groups \cite{Lum2022}, which requires multiple group comparisons. This challenge is resolved by \textit{meta-measures} that aggregate bias measures across all groups into a single measure to evaluate overall model bias. However, when group sizes are unequal, meta-measures can be biased themselves \cite{Lum2022}.

In this paper, we evaluate different performance and bias measures on several End-to-End (E2E) models, before and after using bias mitigation strategies, with an aim to reduce bias in Dutch language against children, teenagers, non-native speakers, and older adults for different speech types. Our goal is to analyze and identify measures that provide insights into performance and bias across diverse speaker groups. We conclude with recommendations for improved reporting of ASR performance and bias which better represent a system’s performance for different speakers groups, and overall system bias.

\section{Background and Related Work}
\subsection{Diversity of Speech}
Human speech is highly variable, and the variability is caused by several factors: speaker's physiology, socio-linguistic background, and speech and language development or proficiency.

The anatomical structure of the speech production organs and their changes due to aging impact~the speech production system, affecting speech characteristics across \textbf{ages} \cite{Fung_genderchild2021}. Children's speech exhibits a higher pitch than that of adults, which drops as their organs develop during their teenage years~\cite{Prakash2015WhatsSI} until they align with those of adults~\cite{ASA_DevelopmentAcoustics_SungbokLee_1999}. In old age, reduced control over vocal organs results in less articulated speech, marked by pauses and slower rate~\cite{ASA_DevelopmentAcoustics_SungbokLee_1999}. Acoustic differences are also influenced by \textbf{gender}, that can vary across languages and speaker origin \cite{pepiot:halshs-00999332}. Generally, female speakers have shorter vocal tract, lighter vocal fold mass, resulting in distinct phonation, lower spectral tilt, and a higher fundamental frequency compared to male speakers \cite{pepiot:halshs-00999332, MENDOZA199659_malevsFemale}. Physiological aspects influencing speech in other genders is under-explored. Speech organ development issues or damage can lead to disordered or \textbf{pathological speech}, deviating from healthy speech patterns. Conditions like dysarthria, caused by facial paralysis or muscle weakness, result in slurred or slow speech which may impact intelligibility~\cite{Dysarthia_disorder}. 

Spoken languages differ across social dimensions (e.g., ethnicity, religion, status, education) and geographical barriers, resulting in diverse \textbf{sociolinguistic} accents and dialects~\cite{SocioLing_2008}. 
Moreover, developmental phases from infancy to adulthood significantly shape speech \textbf{proficiency}~\cite{MilestoneChild_JSLHR}. Speakers proficient in their native language often show a reduced proficiency and native accent when speaking a non-native language~\cite{flege1987production}, which can improve over time and  approach native levels \cite{MilestoneChild_JSLHR}. 

\subsection{Bias in ASR}
The literature on bias in speech recognition can be divided into bias-detecting studies and studies that address or mitigate bias.

\smallskip
\noindent\textit{Bias Detection:} The majority of studies on ASR performance disparities by \textbf{gender} show higher WERs for female speech than for male speech~\cite{Tatman2017, garnerin2019gender, Garnerin2021InvestigatingLibrispeech}. This trend holds across dialects~\cite{Tatman2017}, and is linked with under-representation of female speech in the training data~\cite{Garnerin2021InvestigatingLibrispeech}. Interestingly, a skew towards female speakers does not harm male performance, except when a model is solely trained on female speech. Some studies find no significant performance difference between genders~\cite{Tatman2017EffectsCaptions, Gao2022SeamlessRecognition, YuChan2022TrainingEnglishes}, while others generally found female speech to be better recognized than male speech~\cite{feng2021quantifying, Liu2022TowardsTranscriptions}. 
In \cite{Liu2022Model-basedASR}, WERs for each gender were dependent on the specific test set, but no further analysis was conducted.
\textbf{Age} has a clear impact on bias ~\cite{feng2021quantifying,Gao2022SeamlessRecognition,  YuChan2022TrainingEnglishes, Liu2022TowardsTranscriptions, Patel2023UsingSystems}: ASR models perform worse for children, better for teenagers, followed by older adults, and best for adult speakers \cite{feng2021quantifying,Gao2022SeamlessRecognition,Patel2023UsingSystems}, although \cite{Liu2022TowardsTranscriptions} did not find an age effect.

Studies on \textbf{accents, language proficiency, regional accents} consistently indicate that ASR systems perform worse for non-native speakers~\cite{FENG2024_101567,Patel2023UsingSystems, Zhang2022MitigatingAccents} and regional accents for seen \cite{Tatman2017, Tatman2017EffectsCaptions} and unseen dialects of English  \cite{YuChan2022TrainingEnglishes}, Dutch and Mandarin Chinese \cite{feng2021quantifying, FENG2024_101567}, and French and German \cite{ZiruiThesis}. Notably, \cite{YuChan2022TrainingEnglishes} found that non-native speakers with a non-tonal first language were better recognized than those with a tonal first language, indicating potential bias towards inherent linguistic differences.
Both \cite{FENG2024_101567,garnerin2019gender} found speech type to impact ASR performance with read speech being favored over non-read speech. 
Moreover, performance is worse for speakers of color than for white speakers~\cite{koenecke2020racial, Dheram2022TowardDisparities,Tatman2017EffectsCaptions, Gao2022SeamlessRecognition} (note: \cite{Dheram2022TowardDisparities} used US ZIP codes as a proxy for race), with worse performance for darker skin tones \cite{Liu2022TowardsTranscriptions} showing the impact of \textbf{race and skin types}.

Final notes: Only a few studies consider intersectional demographic groups~\cite{FENG2024_101567, feng2021quantifying, Tatman2017, garnerin2019gender,  Patel2023UsingSystems}. Apart from gender which is typically evaluated as binary (unfortunately), the number and type of groups for a particular factor varies across studies, which makes it hard to compare across studies. While most studies in bias use meta-data to create speaker groups, \cite{Dheram2022TowardDisparities} used geo-demographic cohorts of speakers and automatically discovered cohorts based on speakers' voice similarity. Implicit fairness assumptions typically involve error rate parity, explicitly stated as in~\cite{Liu2022Model-basedASR}, while \cite{Gao2022SeamlessRecognition} assumes fairness as equality of opportunity. 


\smallskip
\noindent\textit{Bias Mitigation:} Research on bias mitigation is scarce; however, the first results are promising. The ``Casual Conversations (CC)" dataset, an 846 h corpus with diverse metadata on gender, age, and skin tone was developed to address data issues in model evaluation  \cite{Liu2022TowardsTranscriptions}. A state-of-the-art ASR model trained on Librispeech with a 2.4\% WER on the Librispeech test-clean gave a 34.3\% WER across 12 speaker groups of the CC dataset. This was reduced to only a 7.2\% WER for a model trained with supervised/semi-supervised learning with up to 2 million h diverse speech, followed by fine-tuning. 

In \cite{Zhang2022MitigatingAccents}, the bias against non-native accents was reduced in an E2E model using speed perturbations, cross-lingual voice conversion-based augmentations, and domain adversarial training, while \cite{zhang22n_s4g} reduced bias against non-native speakers in a hybrid model using multi-task learning and pitch shift for data augmentation. In \cite{Dheram2022TowardDisparities} semi-supervised learning helped mitigate the performance disparities between the best and worst performing speaker groups. Bias against child speech can be reduced by using vocal tract based normalization \cite{Patel2023UsingSystems}. The work in \cite{Zhang2023ExploringDA} showed that data augmentation applied to source speech also improved performance and bias on diverse, target speech, showing that increasing the variability in the standard speech training data improves recognition performance on diverse speech. 
\begin{table}[t]
\caption{Details of the CGN and Jasmin-CGN test sets.}
\vspace{-0.1cm}
\label{Table:CGN|Jasmin}
\centering
\resizebox{\linewidth}{!}{
    \begin{tabular}{cccc|rr}
        \toprule
        \textbf{Datasets} & \textbf{Style} & \textbf{Age} & \textbf{\#Spks}& \multicolumn{2}{c}{\textbf{\#Utterances $|$ \#Hours}} \\
        \cmidrule{2-6}
        &  & & &  Test-Read & Test-CTS/HMI \\ 
        \midrule
        
       CGN         & Read $|$ CTS & 18-65  & 2897 &  409 $|$ 0.45 & 3884 $|$ 1.80 \\
       Jasmin-DC  & Read $|$ HMI & 06-13   & 71 & 13104 $|$ 6.55 & 3955 $|$ 1.55\\
       Jasmin-DT  & Read $|$ HMI & 12-18  & 63 & 9061 $|$ 4.90 & 2723 $|$ 0.94\\
       Jasmin-NnT & Read $|$ HMI & 11-18  & 53 & 11545 $|$ 6.03 & 3093 $|$ 1.16\\
       Jasmin-NnA & Read $|$ HMI & 19-55  & 45 & 11304 $|$ 6.01 & 6808 $|$ 3.07\\
       Jasmin-DOA
       & Read $|$ HMI & 65+ & 68 & 11978 $|$ 6.38 & 8779 $|$ 3.89\\
        \bottomrule
    \end{tabular}}
    \vspace{-0.5cm}
\end{table}

\section{Methodology}

\subsection{The Dutch Corpora}
We use the Corpus Gesproken Nederlands (CGN) \cite{oostdijk2000spoken}, which consists of speech spoken by Dutch adult, native speakers. The type of speech data includes lecture recordings, broadcast news, and spontaneous conversations. The training data is $\sim$430 h; the test sets consist of CGN  read (Rd) broadcast news and Conversational Telephone Speech (CTS). 
The Jasmin corpus~\cite{cucchiarini2006jasmin}, consisting of Read speech and Human Machine Interaction (HMI) speech spoken by native and non-native Dutch-speaking children, teenagers, adults and older adults is used for evaluation. Table \ref{Table:CGN|Jasmin} shows details of the speaker groups, age range in years, and the hours of speech in the CGN and Jasmin test sets.

\subsection{ASR Models and Experimental Setup}
We carry out experiments with two different E2E models and two bias mitigation approaches. In our first experiment, a conformer model \cite{ConformerAnmolG} is trained from scratch without augmentations ($NoAug$). 
The front-end features include $80$ dimensional log-Mel filter-bank features with $3$-dimensional pitch features for network training. Unigram model with 5k byte pair tokens is used. To address and mitigate bias, the conformer model is then trained with speed perturbed speech (SP; at 90\% and 110\% of the original speed rate data ($SpAug$)), and with both SP and spectral augmentation ($SpSpecAug$) applied to the CGN training set. In the second experiment, we compare the performance of the OpenAI-Whisper small model ($Ws$)~\cite{radford2023robust} before and after fine-tuning on the CGN training data ($WsFT_{cgn}$).

All audio files are sampled at 16kHz, and ESPNet \cite{DBLP:ESPnet} is used for the experiments. For both scratch training and fine-tuning, the model is trained for 20 epochs, and the final model is averaged over the last 5 epochs. After decoding, all non-speech tags and symbols are removed from the text file.



\begin{table*}[t]
\caption{Results in \%WER for the E2E Dutch ASR systems when trained on CGN and tested on CGN and Jasmin.}
\vspace{-0.2cm}
\label{Table:DutchResults}
\resizebox{\linewidth}{!}{
\begin{tabular}{@{}cc|cc|ccccc|ccccc|ccc@{}}
\hline

\textbf{}        & \textbf{}     & \multicolumn{2}{c|}{\textbf{CGN}} & \multicolumn{5}{c}{\textbf{Jasmin: Read}}   & \multicolumn{5}{c}{\textbf{Jasmin: HMI}} & \multicolumn{3}{c}{\textbf{Jasmin: Average}} \\ \hline

\textbf{Model}  & \textbf{Approach} & \textbf{Rd}    & \textbf{CTS} & \textbf{DC} & \textbf{DT} & \textbf{NnT} & \textbf{NnA} & \textbf{DOA} & \textbf{DC} & \textbf{DT} & \textbf{NnT} & \textbf{NnA} & \textbf{DOA} &\textbf{Read} & \textbf{HMI} &\textbf{All}\\
\hline

       
       
%

 & $NoAug$  & 9.5 & 23.7      &  41.6& 21.5	& 52.2	& \textbf{57.2}	& 27.7		&	48.7 &	38.6 &	58.2	& 59.4 & 40.9 & 40.0 & 49.2 & 44.6 \\
       
Conformer & $SpAug$   & 6.9 & 21.8 & 35.1	& 19.9	 & 52.8	& 58.5	& 26.7		& 41.4	&32.9	&56.5	&58.4	&39.8	& 38.6 & 45.8 & 42.2	\\
       
 & $SpSpecAug$  & \textbf{6.9} & \textbf{19.9} & \textbf{35.0}	& \textbf{18.4} & \textbf{49.6}	&57.5	&\textbf{25.7}		&\textbf{38.4}	&\textbf{26.4}	&\textbf{50.5}	&\textbf{54.4}	&\textbf{37.2}	& \textbf{37.2} & \textbf{41.4} & \textbf{39.3	} \\
\cmidrule{2-17}
Whisper & $Ws$  & 17.1 &  54.1   & 40.3  & 25.5 & 53.8 & 58.1 & 34.1 &  54.5  & 40.6        & 59.3 & 73.1 & 50.9 & 42.9 & 55.9 & 49.0   \\
       
&  $WsFT_{cgn}$  & 7.8 & 22.7 & 40.9 & 22.4  & 57.7 & 60.8 & 28.2 & 43.5  & 37.6  & 59.6 & 58.3  & 43.92 & 42.3 & 48.6 & 45.8 \\

\bottomrule
\end{tabular}}
\vspace{-0.4cm}
\end{table*}

\begin{table}[t]

\footnotesize
\caption{Different bias measures used in this study.}
\vspace{-0.2cm}

\label{tab:bias_measures}
    \centering
    \resizebox{\linewidth}{!}{
    \begin{tabular}{lcc}
    \toprule
    \textbf{Bias Measures} & 
    \textbf{Equation} & \textbf{Reference}\\ 
    \midrule
    Group-to-min: $G2min_{diff}$ & 
    $\mathnormal{ b_{spk_{g}} - b_{min} }$ & \cite{feng2021quantifying, Zhang2022MitigatingAccents,zhang22n_s4g} \\
    
    Group-to-norm: $G2norm_{diff}$ & 
    $\mathnormal{ b_{spk_{g}} - b_{norm} }$ & \cite{Patel2023UsingSystems}   \\
    \cmidrule{1-3}
    Group-to-min: $G2min_{reldiff}$  & 
    $\mathnormal{(b_{spk_{g}} - b_{min})}/b_{min}$ & \cite{Dheram2022TowardDisparities, Liu2022TowardsTranscriptions} \\
    
    Group-to-norm: $G2norm_{reldiff}$  & 
    $\mathnormal{ (b_{spk_{g}} - b_{norm})}/b_{norm}$ & - \\
    \bottomrule
    \end{tabular}}
    \vspace{-0.5cm}
\end{table}

\subsection{Performance and Bias Measures}
\textit{Performance Measures:} In addition to the standard mean or average ($Avg$) of WER, we compute the median ($Md$), standard deviation ($Stdev$), and range ($Rg$) of the ASR performance. 

\smallskip
\noindent \textit{Bias Measures:} Table~\ref{tab:bias_measures} defines the bias measures from the literature that we compare in this study. They use WER as a base metric ($b$) and calculate the difference (and relative difference) between the WER of a diverse speaker group $spk_{g}$ and a reference speaker group, i.e., the speaker group with the minimum WER ($G2_{min}$) or the norm speaker group ($G2_{norm}$). 

As a meta-measure we use Overall Bias \cite{Patel2023UsingSystems}, which averages the biases across different speaker groups. We compute the difference (Eq. 1) and relative difference (Eq. 2) as follows:
\begin{align}
    \mathnormal{Overall\ Bias_{diff} = \frac{1}{G} \sum_{g} (b_{spk_{g}}-b_{(min,norm)})}.\\
    \mathnormal{Overall\ Bias_{reldiff}  = \frac{1}{G}\sum_{g} \dfrac{(b_{spk_{g}}-b_{(min,norm)})}{b_{(min,norm)}}}.
\end{align} 
where, $b_{min}$ and $b_{norm}$ is the group with minimum WER and WER of the norm group, respectively, and $G$ is the total groups.

\section{Experimental Results}
In this section, we first evaluate our models in terms of recognition performance (Sec. 4.1), followed by bias measures evaluation (Sec. 4.2) and recommendations (Sec. 4.3).

\subsection{ASR System Performance} 
Table~\ref{Table:DutchResults} presents the WER results for the Dutch ASR system, trained on CGN (adult speech) and tested on various speaker groups. The $NoAug$ system achieves 9.5\% and 23.7\% WER on read and continuous speech from CGN adult speakers (matched condition), which is close to state-of-the-art performance on CGN with a language model (which we do not use)~\cite{FENG2024_101567}. On diverse speech, however, performance is much worse, especially for non-native adults, teens, and native children. Speed perturbations ($SpAug$) improved performance for norm and diverse native speaker groups, with the largest improvement for DC, DT, and DOA, but not for non-native speakers. Adding  SpecAug  improved performance for most speaker groups, but mostly for non-native speakers. 
Whisper showed the poorest performance on diverse speech, especially for the non-native speakers. Fine-tuning on CGN only improved performance for read speech by teenager and older adults, while for HMI speech, improvements were observed for all speaker groups except the non-native teenager speech, but falling short of even $SpAug$.

Table~\ref{Table:DutchResultsMeanStdevRange} shows the $Md$, $Stdev$ and $Rg$ of WER over all groups for Read and HMI speech. $SpSpecAug$ has the lowest $Avg$ and lowest $Md$  for both read and HMI speech, although the $Stdev$ is relatively high showing the variability of model performance across the speaker groups (in line with the range of WERs). All models had lower $Avg$ and $Md$ values for Read compared to HMI, however, a higher $Stdev$ for Read than HMI. This is because the bias mitigation approaches used read speech from native speakers, improving the performance of read speech and native groups at the expense of others, resulting in larger variations and bias that were captured by $Stdev$ of the WERs. 



\begin{table}[t]
\centering
\caption{Median, standard deviation and range of WER for the Dutch ASR systems, for Jasmin Read and HMI speech.}
\vspace{-0.3cm}
\label{Table:DutchResultsMeanStdevRange}
\resizebox{1.0\linewidth}{!}{
\begin{tabular}{@{}c|cc|cc|cc@{}}
\toprule
 & \multicolumn{2}{c}{\textbf{Md}} & \multicolumn{2}{c}{\textbf{Stdev}}  & \multicolumn{2}{c}{\textbf{Rg: Max-Min}}  \\ 
\cmidrule{2-7}
\multicolumn{1}{c|}{\textbf{Jasmin}} & \textbf{Read} & \textbf{HMI} & \textbf{Read} & \textbf{HMI} & \textbf{Read} & \textbf{HMI} \\
\midrule
$NoAug$ &41.6	&48.7	&\textbf{15.3}	&\textbf{9.6} & 57.2-21.5 & 59.4-38.6 \\
$SpAug$ &35.1	&41.4	&16.6&11.1 & 58.2-19.9 & 58.4-32.9 \\
$SpSpecAug$ &\textbf{35.0}	&\textbf{38.4}	&16.3	&11.2 & 57.5-18.4 & 54.4-26.4 \\
\midrule
$Ws$ & 40.3 & 54.5 & 15.6 & 13.7 & 58.1-25.5 & 73.1-40.6\\
$WsFT_{cgn}$ & 40.9 & 43.9& 19.8 & 9.8 & 60.8-22.4 & 58.3-37.6\\

\bottomrule
\end{tabular}
}
\vspace{-0.6cm}
\end{table}

\begin{figure*}[t!] 
  \centering
  \includegraphics[trim={0cm 0cm 0cm 1.0cm}, clip, width=\linewidth]{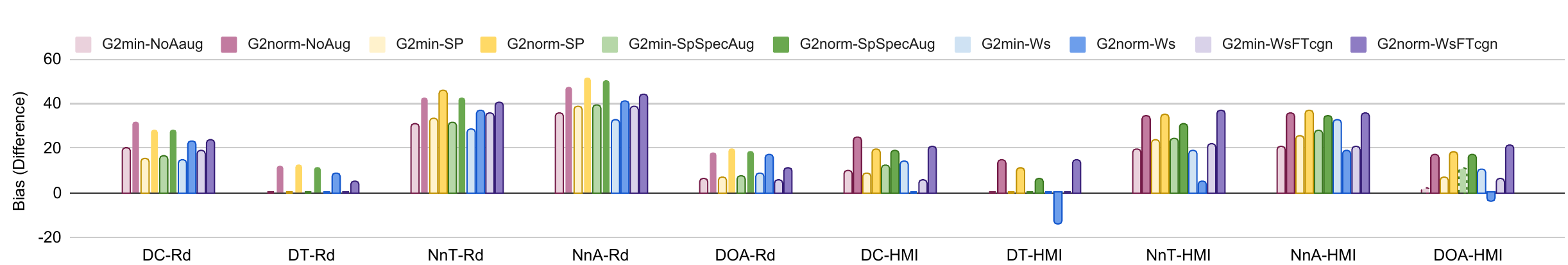} 
  \vspace{-0.2cm}
   \includegraphics[trim={0cm 0.1cm 0cm 0.3cm}, clip, width=\linewidth]{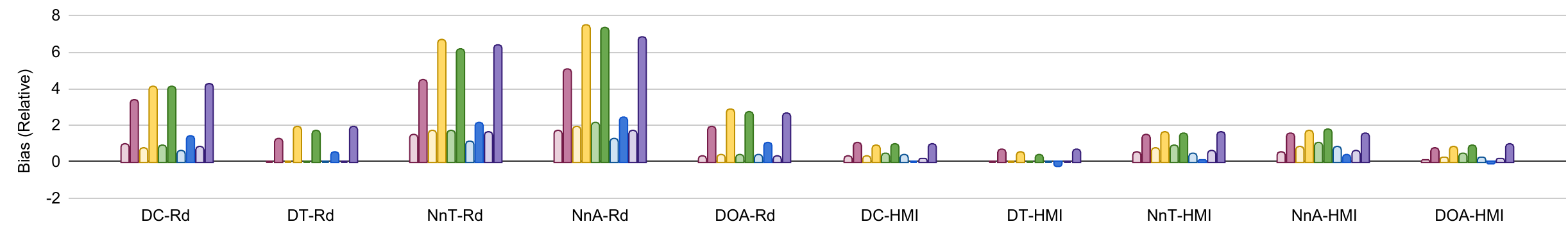} 
  \caption{Bias estimated for the Read and HMI test sets  using the difference measure (top panel) and relative difference (bottom panel) for both G2min and G2norm using the three conformer models and two Whisper models.}
  \label{Fig:DutchResultsBias_Diff_Rel_both}
  \vspace{-0.4cm}
\end{figure*}

\subsection{Evaluating the Bias Measures} 

\textit{Group-based Bias}: Figure \ref{Fig:DutchResultsBias_Diff_Rel_both} shows the bias computed using the difference (top panel) and relative difference (bottom panel) measures for the ASR systems in Table \ref{Table:DutchResults}. For $G2_{min}$, the min.~WER group is DT, while for $G2_{norm}$, the norm group is the CGN test-set. Bias is largest against the non-natives, in line with the WER results in Table \ref{Table:DutchResults}. Bias decreases for DC, DOA and is the smallest for DT (which has the closest acoustic match to the training data). HMI speech shows lower bias than read speech, despite higher WERs for HMI, this is because the reference group has a higher WER for HMI speech than read speech.
While augmentations should ideally enhance performance and reduce bias, the $Sp$ and $SpSpecAug$ models, despite improving average WER, mostly show increased bias against non-native speakers. Similarly, fine-tuning, while often shown to reduce WER, increases the bias compared to $Ws$. For $WsFT_{cgn}$, this increase in bias is more pronounced for $G2_{norm}$ compared to $G2_{min}$ and also more visible in read speech than in HMI, because the norm group from read speech has a lower WER after fine-tuning. 
With relative difference, $G2_{min}$ and $G2_{norm}$ show similar patterns of increase and decrease in bias. However, for the difference-based measures, the patterns might be different, emphasizing the need for nuanced bias evaluation measures. 

\begin{figure}[h]  
\vspace{-0.3cm}
  \centering
  \includegraphics[trim={0.2cm 0.0cm 0cm 0cm}, clip, width=\linewidth]{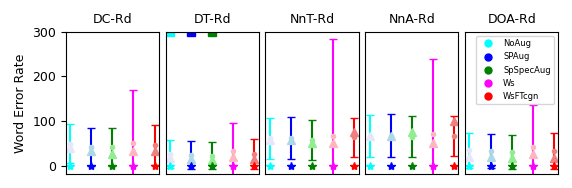} 
  \includegraphics[trim={0.2cm 0.3cm 0cm 0.1cm}, clip, width=\linewidth]{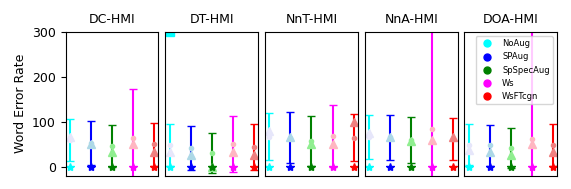} 
  \caption{Performance Measures: $Avg \pm Stdev$ ($\bullet$), $Md$ (\scalebox{0.8}{$\blacktriangle$}), $Rg$:min ($\star$) and max (\scalebox{0.5}{$\blacksquare$}) for the individual speaker groups; (top): Read speech, (bottom): HMI speech, for all models.}
  \label{Fig:WithinSpkAllsystem}
  \vspace{-0.2cm}
\end{figure}

Figure \ref{Fig:WithinSpkAllsystem} shows the $Avg$$\pm$$Stdev$ (vertical lines), $Md$, Min and Max (=$Rg$) computed \textit{within} the different speaker groups. First, all systems have a minimum WER of 0, so for each speaker group and each model there are utterances that are entirely correctly recognized. The maximum WER is very high across models for the different speaker groups, indicating that some utterances that are very poorly decoded.
Mostly across diverse speaker groups and speech types, a consistent pattern emerges: the $Md$ tends to be lower than $Avg$. 
This shows an uneven distribution of WER values even \textit{within} a speaker group, and provides valuable input for mitigating bias.
The $Stdev$ are high for non-natives, followed by DC, DOA and DT, which correlates with the bias measures shown in Figure \ref{Fig:DutchResultsBias_Diff_Rel_both}.
The $Stdev$ values are higher for $Ws$ than the other models indicating larger WER variations even within speaker groups (and least for DT), $Ws$ is the only model that was not trained on CGN, leading to a training/test set mismatch. This high variance is reduced for $WsFT_{cgn}$ and its $Stdev$ is similar to the conformer models.

\noindent \textit{System Bias}: The Overall Bias in Table \ref{Table:DutchResultsOverallBias_ReadHMI} shows that the bias for HMI is lower than for Read speech, aligning with the high $Stdev$ of Read speech compared to HMI speech (Table \ref{Table:DutchResultsMeanStdevRange}). Overall, $Ws$ exhibits the lowest bias across all measures (despite high variance in WER within speaker groups), although this is not entirely consistent with the $Stdev$  reported in Table \ref{Table:DutchResultsMeanStdevRange}. This is due to $Stdev$ being measured with respect to mean WER, while bias is measured with respect to the group with the minimum WER or the norm group. Also the overall bias measures capture the earlier findings that the mitigation approaches do not reduce bias despite improving performance. $G2_{min}$ and $G2_{norm}$ do not show similar trends and hence, do not correlate well with each other, especially when only difference is used.



\subsection{Recommendations}
\textit{Performance Measures:} 
Avoid overemphasizing averaged results; they can hide disparities in performance among specific demographic groups. The mean being a numerical average, can be sensitive to extreme values, whereas the median can offer a robust measure of central tendency. While the range is useful for between-speaker group comparisons, it may not always be suitable for within-speaker group comparisons due to outliers (easily recognized speech files, non-speech utterances, short or long files in the test set). $Stdev$, on the other hand, captures performance variations between and within speaker groups,  and can be used to make decisions on the trade-offs between achieving the best average performance and maintaining a fair system. 

\smallskip
\noindent \textit{Bias Measures:} Using $G2_{min}$ has limitations, as shown in Figure \ref{Fig:DutchResultsBias_Diff_Rel_both}, where bias estimation for the minimum WER error group is 0, providing no clarity about its bias. Likewise, $G2_{norm}$ may yield negative values, as observed with $Ws$ when the WER of the diverse group is lower than the norm group. Using $G2_{norm}$ is only feasible when the training data is available. When the training data is not accessible, such as with pre-trained models, using $G2_{min}$ is preferred. 
Comparison of the difference and relative difference measures shows that, while a direct difference measures the magnitude of disparity, suitable for straightforward comparisons, relative difference expresses differences as percentages or proportions relative to one of the values, especially when dealing with values of varying scales. 



To comprehensively capture bias within and across different speaker groups, we recommend 
to  break down error rates to demographic groups and variables like speech type. The ASR performance should be reported using the median and standard deviation, and bias using a relative measure that considers all speaker groups (both diverse and standard) and then use the group with the minimum WER as a reference. A final note, unlike WER, there is no ground-truth in bias (only human judgement), hence, user feedback and experiences should be considered for qualitative insights into the real-world impact of bias.

\begin{table}[t]
\caption{Overall bias estimated using the difference and relative difference for Read and HMI speech.} 
\vspace{-0.2cm}
\label{Table:DutchResultsOverallBias_ReadHMI}
\resizebox{\linewidth}{!}{
\begin{tabular}{@{}ccc|cc|cc|cc@{}}
\toprule

& \multicolumn{2}{c}{\textbf{$G2min_{\textbf{\textit{diff}}}$}} & \multicolumn{2}{c}{\textbf{$G2norm_{\textbf{\textit{diff}}}$}} & \multicolumn{2}{c}{\textbf{$G2min_{\textbf{\textit{reldiff}}}$}} & \multicolumn{2}{c}{\textbf{$G2norm_{\textbf{\textit{reldiff}}}$}} \\ 
\cmidrule{2-9}
& \textbf{Read}  & \textbf{HMI} & \textbf{Read} & \textbf{HMI} & \textbf{Read}  & \textbf{HMI} & \textbf{Read} & \textbf{HMI} \\
\midrule
$NoAug$ & 23.18 & 13.20  & 30.54  & 25.46 & 1.08  & 0.34  & 3.21 & 1.08  \\
$SpAug$ & 23.38 & 16.13 & 31.70 & 24.00  & 1.17  & 0.49  & 4.59 & 1.10 \\
$SpSpecAug$ & 23.55 & 18.73 & 30.34  & 21.48 & 1.28  & 0.71  & 4.40  & 1.08  \\
\midrule
$Ws$ & 21.08 & 18.85 & 25.26  & 1.58  & 0.83  & 0.46  & 1.48 & 0.03  \\
$WsFT_{cgn}$ & 24.50  & 13.73 & 24.9 & 25.88 & 1.10 & 0.37  & 4.38 & 1.14  \\ 
\bottomrule
\end{tabular}

}
\vspace{-0.2cm}
\end{table}

\section{Conclusion}
The error rate is a fundamental metric offering a holistic system evaluation, however, our granular, intersectional analysis of  different demographic speaker groups showed that error rate does not reflect the performance and bias within and across speaker groups well. In line with the potential pitfalls, there is a clear need for performance and bias measures to capture performance variation, and this paper gives recommendations on it. 

\newpage

\bibliographystyle{IEEEtran}
\bibliography{Bias-ASR-IS24}

\end{document}